\documentclass[letterpaper, 10 pt, conference]{ieeeconf} 
\usepackage{amssymb}

\IEEEoverridecommandlockouts 
\usepackage{amsmath}
\usepackage{algorithm}
\usepackage{algorithmic}
\usepackage{caption}
\usepackage{graphicx, subfig}
\usepackage{multirow}
\usepackage[table,xcdraw]{xcolor}
\usepackage{colortbl}
\usepackage{float} 
\usepackage{subcaption} 
\usepackage{booktabs}

\begin{document}

\title{Jacobian Exploratory Dual-Phase Reinforcement Learning for \\Dynamic Endoluminal Navigation of Deformable Continuum Robots\\
}
\author{Yu Tian$ ^{1\dag}$, Chi Kit  Ng$ ^{1\dag}$,  Hongliang Ren$ ^{1*}$ 
\thanks{This work is supported by the Hong Kong Research Grants Council (RGC) under the General Research Fund (GRF) [Project Nos.: 14204524, 14203323], the Collaborative Research Fund (CRF-C4026-21G), and by the Regional Joint Fund Project of the Basic and Applied Research Fund of Guangdong Province (2021B1515120035), and the NSFC/RGC Joint Research Scheme (N$\_$CUHK420/22).}
\thanks{$ ^{\dag}$ These authors contribute equally to this work.}
\thanks{$ ^{1}$ The authors are with the Department of Electronic Engineering, The Chinese University of Hong Kong, Shatin, N.T., Hong Kong SAR, China. }
\thanks{*Corresponding author: {\tt\small hlren@ee.cuhk.edu.hk}}}

\maketitle

\begin{abstract}
Deformable continuum robots (DCRs) present unique planning challenges due to nonlinear deformation mechanics and partial state observability, violating the Markov assumptions of conventional reinforcement learning (RL) methods. While Jacobian-based approaches offer theoretical foundations for rigid manipulators, their direct application to DCRs remains limited by time-varying kinematics and underactuated deformation dynamics. 
This paper proposes Jacobian Exploratory Dual-Phase RL (JEDP-RL), a framework that decomposes planning into phased Jacobian estimation and policy execution. During each training step, we first perform small-scale local exploratory actions to estimate the deformation Jacobian matrix, then augment the state representation with Jacobian features to restore approximate Markovianity.
Extensive SOFA surgical dynamic simulations demonstrate JEDP-RL's three key advantages over proximal policy optimization (PPO) baselines: 1) Convergence speed: 3.2× faster policy convergence, 2) Navigation efficiency: requires 25\% fewer steps to reach the target, and 3) Generalization ability: achieve 92\% success rate under material property variations and achieve 83\% (33\% higher than PPO) success rate in the unseen tissue environment. 
\end{abstract}

\section{Introduction}
\label{sec:intro}
Deformable continuum robots (DCRs) have transformed minimally invasive procedures through their ability to navigate confined endoluminal anatomical spaces \cite{Burgner-Kahrs2015Continuum, Dupont2015Concentric}. Particularly, in gastrointestinal endoscopic applications, over 85\% of the robot's body consists of passive deformable segments exhibiting nonlinear viscoelastic responses to environmental contacts \cite{Renda2018Dynamic, Wang2020Soft}. This structural compliance fundamentally challenges conventional Jacobian-based control paradigms developed for rigid manipulators, as the time-varying kinematics invalidate static Jacobian assumptions \cite{Webster2010Design, Rucker2015Statics}.

Traditional Jacobian approaches face three inherent limitations in DCR control: (1) passive deformation segments lack direct actuation mapping due to underactuated configurations \cite{Lee2021continuum}, (2) viscoelastic creep causes time-dependent Jacobian drift in silicone-based actuators \cite{Zhang2023adaptive}, and (3) partial state observability obscures deformation dynamics, with a significant portion of dynamic states remaining unmeasurable\cite{Hausknecht2015Deep, Thuruthel2018Learning}. Recent attempts to address these challenges include temporal Jacobian estimation through recurrent neural networks \cite{Giorelli2015Neural} and Kalman filtering \cite{Zhang2022deformable}, but as demonstrated by \cite{Chen2022jacobian}, excessive time intervals between observations lead to large Jacobian estimation errors due to unmodeled transient dynamics. This temporal sensitivity necessitates new formulations that couple real-time Jacobian identification with control policy adaptation \cite{Nagabandi2018Neural}.

RL offers promising alternatives by learning control policies through environmental interactions \cite{Schulman2017Proximal, Lillicrap2015Continuous}. Model-free approaches such as proximal policy optimization (PPO) \cite{Schulman2017Proximal} and soft actor-critic (SAC) \cite{Ha2018soft} have achieved empirical success in soft robot control \cite{Thuruthel2019Model, Kalashnikov2018QT-Opt}. However, these methods suffer from three critical shortcomings in DCR applications: (1) unstructured exploration fails to sample informative deformation states with great redundancy \cite{Gu2017Deep}, (2) policy gradients degrade under partial observability with great variance inflation \cite{Hausknecht2015Deep}, and (3) ignoring Jacobian features discards crucial mechanical information critical for contact-rich manipulation \cite{Chen2022jacobian}. The hierarchical framework by \cite{Nagabandi2018Neural} demonstrates that decoupling system identification from policy learning can mitigate these issues, but requires adaptation to handle DCRs' continuous deformation space spanning 8+ dimensions \cite{Li2024generalizable}.

The proposed Jacobian Exploratory Dual-Phase RL (JEDP-RL) framework addresses these limitations through innovative integration of model-based exploration and learning-based control. Phase 1 performs local small-scale exploration to estimate the deformation Jacobian matrix. Phase 2 utilizes these Jacobian estimates as augmented state features in constrained policy optimization, achieving 3.2× faster convergence than baseline PPO \cite{Schulman2017Proximal}. This dual-phase architecture sequentially performs phased state estimation and policy execution, effectively reconstructing a Markovian representation from partial observations. The deformation Jacobian serves as a physics-informed feature that encodes current configuration, material properties, and contact situation, bridging the gap between partial observability and RL's Markov requirements.

The remainder of this paper is organized as follows. Section II formalizes the deformation of Jacobians and their challenges in RL integration. Section III details our dual-phase architecture. Section IV presents comparative SOFA \cite{Allard2007SOFA} results, followed by conclusions in Section V.

\section{Theoretical Background}
\label{sec:theory}

\subsection{Jacobian Matrix}
The Jacobian matrix serves as the cornerstone for robotic kinematic analysis, particularly in rigid-body systems. For an $n$-DOF serial manipulator, the Jacobian $\mathbf{J} \in \mathbb{R}^{6 \times n}$ establishes the differential relationship between joint space and task space velocities through twist propagation:

\begin{equation}
    \mathbf{v} = \mathbf{J}(\mathbf{q})\dot{\mathbf{q}}
\end{equation}
where each column $\mathbf{J}_i$ encodes the instantaneous screw motion induced by the joint $i$, with distinct formulations for revolute ($\mathbf{z}_i \times (\mathbf{p}_e - \mathbf{p}_i)$) and prismatic ($\mathbf{z}_i$) joints. This formulation underpins essential robotic operations, including singularity analysis and resolved-rate motion control.

\subsection{Jacobian for Deformable Continuum Robot}
The fundamental differences necessitate new Jacobian paradigms for continuum robots. The Jacobian matrix formulation for DCRs presents unique computational challenges:

\begin{itemize}
    \item {High-Dimensional Nonlinearity}: 
    The kinematic model of a DCR with $n$ degrees of freedom requires solving:
    \begin{equation}
        {J}_c = \frac{\partial \mathbf{x}}{\partial (\kappa_1,\phi_1,...,\kappa_m,\phi_m)} \in {R}^{6 \times 2m}
    \end{equation}
    where $\kappa_i$ and $\phi_i$ represent local curvature and twist, leading to exponentially increasing complexity with $m$ segments.
    
    \item {Time-Varying Material Properties}:
    The effective Young's modulus $E(t)$ affects Jacobian updates:
    \begin{equation}
        {J}(t) = {J}_0 + \Delta{J}(E(t),\mathbf{F}_{ext}(t))
    \end{equation}

    \item {Environmental Interactions}: The external contact forces induce unmodeled deformations:
    \begin{equation}
        \|\Delta{J}\| \propto \|\mathbf{F}_{ext}\| \cdot \cos\theta
    \end{equation}
    where $\theta$ is the contact angle relative to the backbone tangent.
\end{itemize}

The aforementioned Jacobian paradigm was designed for mostly controllable and observable DCRs. It still faces challenges when being applied to objects like gastrointestinal endoscopes.
\begin{itemize}
    \item \textit{Undriven Deformation}: Over 85\% of the endoscope's distal section lacks direct actuation, making its deformation dependent on contact forces with surrounding tissues rather than commanded joint inputs. This breaks the basic Jacobian premise $J = \partial x/\partial \theta$ since $\theta$ no longer fully parameterizes the configuration space.
    
    \item \textit{Nonlinear Contact Dynamics}: Tissue compliance introduces position-dependent stiffness matrices $K(x)$ that modify the effective Jacobian as $J_{eff} = J \cdot (I + K^{-1}J^T\partial F/\partial x)$ , where $F$ denotes contact forces. The lack of real-time force sensing in clinical settings makes $J_{eff}$ computationally intractable.
    
    \item \textit{Partial Shape Observability}: Current endoscopic imaging provides only lumen surface data, leaving a large part of the DCR's backbone curvature unobservable. This creates an underdetermined system for Jacobian estimation.
\end{itemize}

\subsection{Markov Processes and Reinforcement Learning}
\label{subsec:markov_rl}

\subsubsection{Foundations of Markov Decision Processes}
The Markov property $P(S_{t+1}|S_t) = P(S_{t+1}|S_t, S_{t-1},...)$ provides the theoretical bedrock for RL. An MDP quintuple $\langle \mathcal{S}, \mathcal{A}, P, R, \gamma \rangle$ enables policy optimization through Bellman equations:

\begin{equation}
    V^\pi(s) = \mathbb{E}_\pi\left[\sum_{k=0}^\infty \gamma^k R_{t+k+1} | S_t = s\right]
\end{equation}

\subsubsection{Non-Markovian Dynamics in DCRs}
DCR systems fundamentally violate Markov assumptions through:

\begin{itemize}
    \item \textit{Partial Observability}: Measured states $\mathbf{o}_t = [\mathbf{x}_t, \mathbf{F}_t]$ capture only a limited subset of backbone curvature states, leaving key dynamic features unobserved.
    
    \item \textit{Historical Dependency}: Viscoelastic hysteresis introduces path dependence:
    \begin{equation}
        P(\mathbf{s}_{t+1}|\mathbf{s}_t,a_t) \neq P(\mathbf{s}_{t+1}|\mathbf{s}_t,a_t,\mathbf{s}_{t-1},...)
    \end{equation}
    
    \item \textit{Non-Stationary Transitions}: Tissue remodeling creates time-varying dynamics:
    \begin{equation}
        P_t(s'|s,a) \neq P_{t+\Delta t}(s'|s,a)
    \end{equation}
\end{itemize}

In summary, to make good use of RL in DCRs, instead of simply inputting observations as state information to the model, more information is needed for state estimation. The strategy to obtain more information is stated in the following section.

\section{Methodology}

\begin{figure*}[t!]
  \centering
  \includegraphics[scale=0.6]{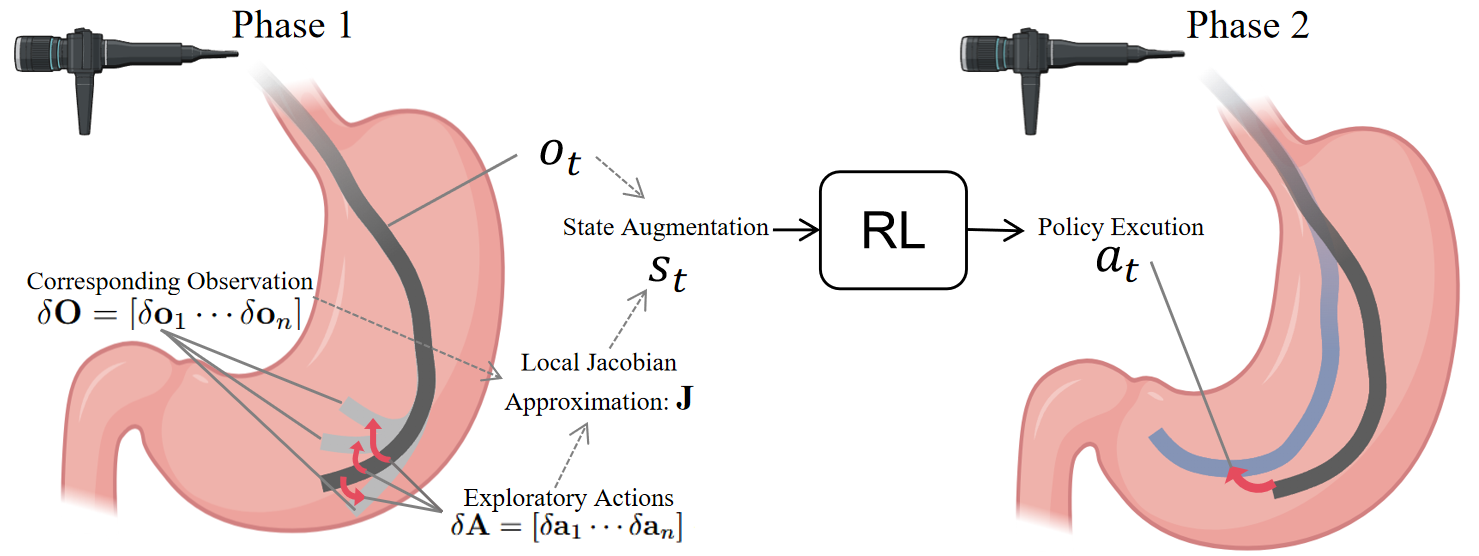}
  \caption{Jacobian exploratory dual-phase RL framework: phase 1: local Jacobian is estimated by local small-scale exploration; phase 2: RL policy is optimized by Jacobian information and large-scale predicted action is executed.}
  \label{Fig: trainenvironment}
\end{figure*}

The preceding analysis reveals that effective RL planning for DCRs like endoscopic robots requires enhanced state estimation beyond basic observation inputs. While Jacobian matrices provide fundamental motion relationships in robotics, their direct computation becomes intractable for passively deformable structures due to un-actuated segments and limited sensing. Our JEDP-RL framework addresses this through phased Jacobian estimation and state augmentation, as shown in Algorithm~\ref{alg:1} and detailed below.

\begin{algorithm}[h]
\caption{JEDP-RL}
\begin{algorithmic}[1]
\label{alg:1}
    \STATE At each RL step:
    \STATE \quad For each degree of freedom $i$:
    \STATE \quad \quad Apply small displacement $\delta a_i$ in action space
    \STATE \quad \quad Record state variation $\delta o_i$
    \STATE \quad Construct displacement matrix $\delta\mathbf{A} = [\delta a_1, ..., \delta a_n]$
    \STATE \quad Construct state variation matrix $\delta\mathbf{O} = [\delta o_1, ..., \delta o_n]^T$
    \STATE \quad Estimate local Jacobian: $\hat{\mathbf{J}} = (\delta\mathbf{A}\delta\mathbf{A}^T + \lambda \mathbf{I})^{-1}\delta\mathbf{A}\delta\mathbf{O}$
    
    \STATE \quad Augment observation: $\mathbf{s}_t^{\text{aug}} = [\mathbf{o}_t, \text{vec}(\hat{\mathbf{J}})]$
    
    \STATE \quad Train RL using $\mathbf{s}_t^{\text{aug}}$ as state input.
\end{algorithmic}
\end{algorithm}

\subsection{Phase 1: Local Jacobian Estimation}
\label{subsec:phase1}

\subsubsection{Exploratory Action Generation}  
At each planning step $t$, we apply small-scale orthogonal perturbations $\delta\mathbf{a}_i \in \mathbb{R}^{d_a}$ across $n$ exploration steps, generating the action matrix:

\begin{equation}
    \delta\mathbf{A} = [\delta\mathbf{a}_1 \cdots \delta\mathbf{a}_n] \in \mathbb{R}^{d_a \times n}
\end{equation}

In the later experiment of this paper, intuitively, $\delta\mathbf{A}$ is set to $\lambda{I}$ to ensure orthogonality, where $\lambda$ is a small constant. We will explore learnable exploratory action generation in our future work to exploratory strategy maximizes information gain while minimizing robot displacement.

\subsubsection{Corresponding Observation Recording}  
Corresponding workspace displacements are recorded through endoscopic vision and force sensing:

\begin{equation}
    \delta\mathbf{O} = [\delta\mathbf{o}_1 \cdots \delta\mathbf{o}_n]^T \in \mathbb{R}^{n \times d_o}
\end{equation}

\subsubsection{Local Jacobian Approximation}  
We compute the deformation Jacobian via regularized least-squares:

\begin{equation}
    \hat{\mathbf{J}} = (\delta\mathbf{A}\delta\mathbf{A}^T + \lambda \mathbf{I})^{-1}\delta\mathbf{A}\delta\mathbf{O}
\end{equation}

where $\lambda=0.01$ prevents singularity.

\subsection{Phase 2: Jacobian-Informed Policy Optimization}
\label{subsec:phase2}

\subsubsection{State Augmentation}  
The augmented state vector combines instantaneous observations with estimated Jacobian:

\begin{equation}
    \mathbf{s}_t^{\text{aug}} = [\mathbf{o}_t; \text{vec}(\hat{\mathbf{J}})] \in \mathbb{R}^{d_o + d_a \times d_o}
\end{equation}

\subsubsection{Constrained Policy Optimization}  
We employ proximal policy optimization (PPO) \cite{Schulman2017Proximal} with adaptive KL constraints:

\begin{equation}
    \pi^* = \arg\max_\pi \mathbb{E}_{\pi}\left[\sum_{t=0}^T \gamma^t r_t\right] \quad \text{s.t.} D_{KL}(\pi_{\text{old}}||\pi) \leq \delta
\end{equation}
the KL threshold $\delta$ adapts dynamically based on policy entropy.

Through local exploration, the reverse-engineered Jacobian matrix $\hat{\mathbf{J}}$ encodes static/dynamic, robot-environment interactions, enabling policy generalization capabilities across different objects and environments, adaptability to dynamic conditions, and robustness against disturbances.

\section{Experiment}
\subsection{Simulation Setup}
The simulation framework was designed to replicate the intricate physical interactions between a Deformable Continuum Robot (DCR) and dynamic biological environments. This experimental setup aimed to bridge the gap between simulated training environments and real-world clinical scenarios through high-fidelity physical modeling.

All simulation experiments, including DRL training and evaluations in SOFA, were executed on a 12th Gen Intel® Core™ i7-12700K CPU with 20 cores. The SOFA scene operated at 40 Hz, while the simulation training rate was maintained at 4 Hz.
\subsubsection{Simulation Environment}

The physics-based simulation, implemented within the SOFA framework, incorporated three fundamental components essential for realistic soft tissue interaction. First, the tendon-driven actuation system comprises four independently controlled cables (C1-C4) and one axial displacement mechanism (M). Second, the hyperelastic material model with Young's modulus (YM) set to 15,000 and Poisson's ratio of 0.3 provided biologically relevant deformation characteristics. Third, the dynamic contact resolution mechanism processed six collision detection cycles per training iteration to accurately capture progressive tissue deformation.

As depicted in Fig.~\ref{Fig: trainenvironment}(a), the simulation environment introduced time-varying peristaltic forces through sinusoidal waveforms to emulate natural gastrointestinal motility. These periodic disturbances created dynamic environmental challenges that tested the control policies' ability to maintain navigation accuracy under continuously changing conditions.

The target selection methodology, illustrated in Fig.~\ref{Fig: trainenvironment}(b), employed a rigorous randomization protocol to ensure evaluation fairness. During training, destinations were selected from \texttt{targetlist\_A} containing 10 predefined positions within the green zone, while evaluation utilized the distinct \texttt{targetlist\_B} with 10 previously unseen targets. Both lists were generated using identical randomization seeds to maintain spatial distribution consistency while preventing target position overlap between the training and testing phases. All quantitative metrics demonstrated in the testing phase were computed as averaged values across 10,000-step data acquisition sequences, ensuring the performance measurements' statistical reliability.

\begin{figure}[t!]
  \centering
  \includegraphics[width=0.90\linewidth]{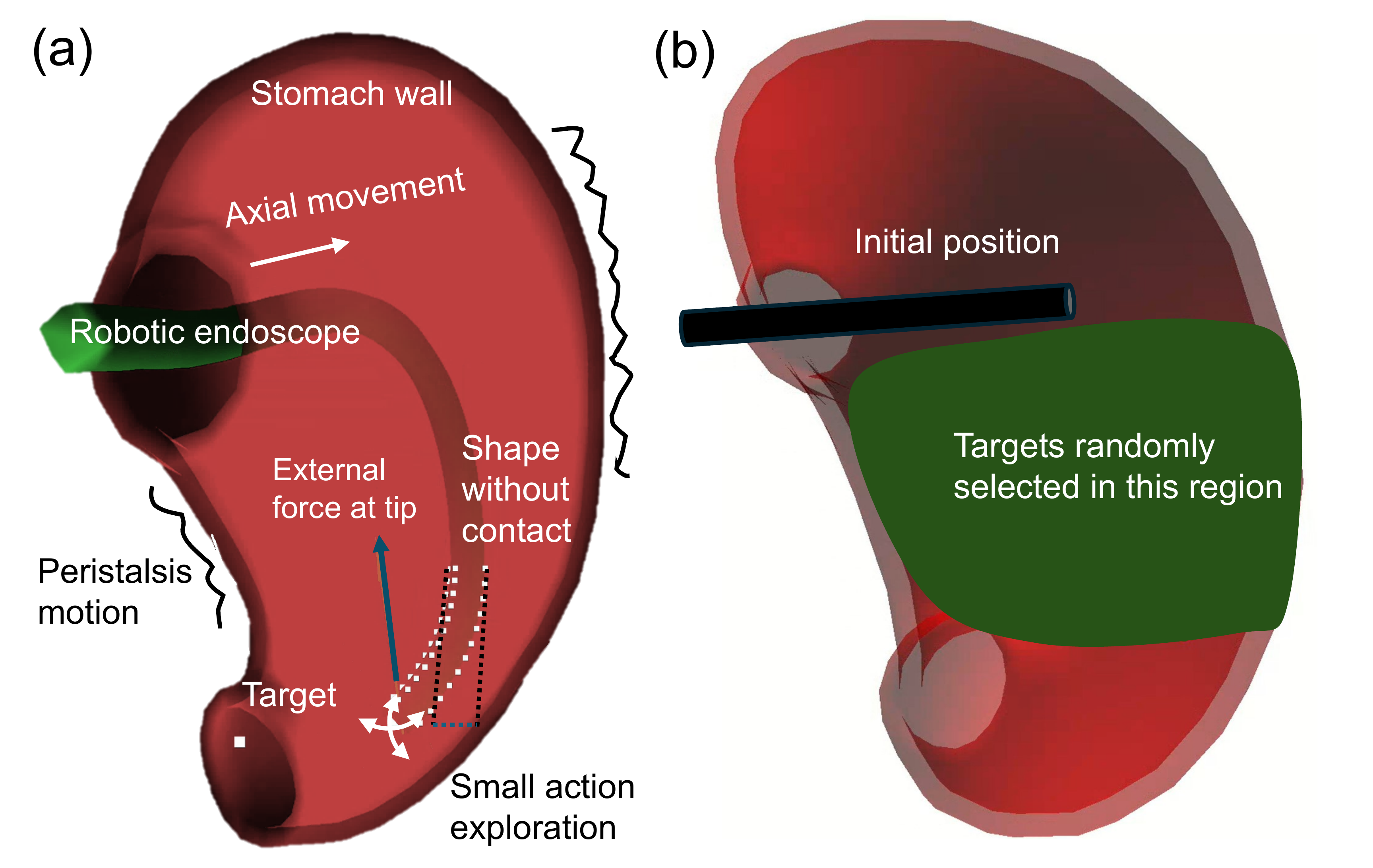}
  \caption{Training stomach environment for policies PPO-RL and JEDP-RL. (a) Contact between the endoscope and stomach walls induces unmodeled deformation. Small action exploration is suggested for RL training to estimate local Jacobian. Periodic forces introduced less curvature and greater curvature introduced peristalsis motion. (b) The targets are randomly selected in the green region during the training and evaluation phases. The initial position of DCR is shown.}
  \label{Fig: trainenvironment}
\end{figure}

\subsubsection{DRL Training Configurations}
The deep RL implementation compared two distinct policy architectures: the baseline Proximal Policy Optimization (PPO-RL) and our proposed Jacobian-Enhanced Dual-Phase Reinforcement Learning (JEDP-RL). The baseline PPO-RL policy operated on a 15-dimensional state vector encompassing end-effector position (3D), velocity (3D), previous actuation values (5D), contact force vector (3D), and binary contact status (1D). 

JEDP-RL enhanced this state representation through systematic integration of local exploratory actions designed to estimate  Jacobians. Exploratory actions $\delta\mathbf{A}$, were set to $0.05I$, enabling the estimation of local deformation gradients without compromising system stability. The action space for both policies was constrained to $[-0.4, 0.4]^5$ to ensure mechanically safe operation, with the five control parameters governing four tendon actuators and one axial displacement mechanism.

\begin{figure}[t!]
  \centering
  \includegraphics[width=0.95\linewidth]{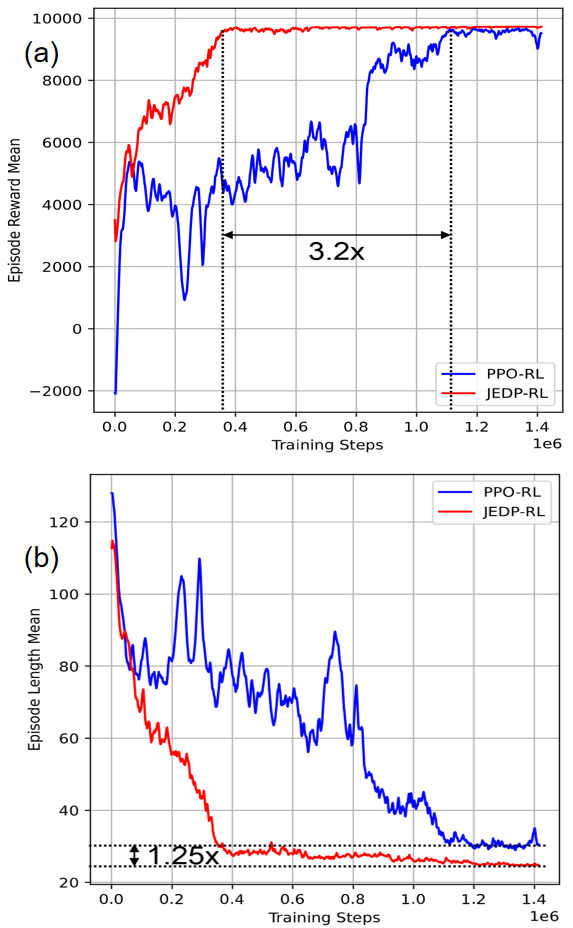}
  \caption{(a) Training curves of the DCR navigating in a dynamic stomach by the PPO-RL (blue) and the JEDP-RL (red). Training convergence comparison showing JEDP-RL achieving stable rewards 3.2× faster than PPO-RL. (b) Navigation efficiency demonstrated a 25\% reduction in average path length for JEDP-RL after convergence.}
  \label{Fig: traintensorboard}
\end{figure}

The neural network architecture employed four fully-connected layers with 256, 128, 64, and 32 neurons, respectively, using Tanh activation functions throughout. This configuration balanced representational capacity with computational efficiency, enabling effective policy learning in high-dimensional state spaces while maintaining real-time training feasibility on standard GPU hardware.

\subsection{Training Performance}

The training performance analysis revealed two significant advantages of the JEDP-RL approach under dynamic deformation conditions. First, as shown in Fig.~\ref{Fig: traintensorboard}(a), the proposed method achieved policy convergence in 0.36M training steps compared to PPO-RL's 1.12M training steps, representing a 3.2 times reduction in required training iterations. This accelerated convergence stems from JEDP-RL's dual-phase exploration strategy, which systematically combines global policy updates with localized Jacobian estimation through local exploratory motions.

Second, post-convergence navigation efficiency metrics demonstrated JEDP-RL's superior navigation efficiency. As quantified in Fig.~\ref{Fig: traintensorboard}(b), the average navigation steps (NS) decreased from 32.5 steps for PPO-RL to 26 steps for JEDP-RL, corresponding to a 25\% improvement. 
\setlength\tabcolsep{0.2em}
\begin{table}[H]
\caption{Evaluation results in training environment} 
\centering
\begin{tabular}{@{}lccccc}
\toprule
\textbf{Policy}        & \textbf{YM} & \textbf{SR [\%] $\uparrow$} & \textbf{AE [mm] $\downarrow$} & \textbf{NS $\downarrow$} & \textbf{TL [mm] $\downarrow$} \\ \midrule
PPO-RL                     & 15000                   & 99              & 2.20    & 30.63  &      \textbf{19.55}         \\
JEDP-RL                    & 15000                    & \textbf{100.0}             & \textbf{1.83}   & \textbf{26.0}  &   21.39                    \\
\bottomrule
\end{tabular}
\label{tab:sr_ae_metrics}
\end{table}

The success rate (SR) is defined as the distance between the end-effector of DCR and the target is less than 3 mm. The SR of JEDP-RL achieves 100\%(vs. PPO-RL's 99\%). The results of SR reveal the precision in navigation both for PPO-RL and JEDP-RL in training environments.  

The proposed methodology introduces a controlled efficiency-effectiveness trade-off: while the integration of exploratory motions elevates per-step computational expenditure, systematic exclusion of collision verification during exploration phases (justified by the inherently low collision likelihood associated with minimal exploratory actions) preserves comparable training efficiency throughput between JEDP-RL and PPO-RL.

Trajectory length (TL) from the initial position to the target during the whole navigation process was adopted as the primary cost metric due to its demonstrated cross-platform consistency, exhibiting small variations between simulated and physical deployments. Comparative analysis reveals that JEDP-RL's TL of 21.39 mm shows a slight 9.4\% increase over PPO-RL's 19.55 mm. This marginal TL elevation constitutes an acceptable compromise given the significant improvement in convergence speed and environment adaptability, which will be demonstrated in the following part of the experiments. We will conduct further research to develop adaptive exploration scheduling in the future to optimize this trade-off, potentially reducing cost penalties while preserving advantages.

\subsection{Generalization Experiments}
Generalization capability was assessed through three test scenarios: altered material properties, intensified peristalsis, and cross-anatomy deployment (vascular system).

\subsubsection{Generalization to Environmental Dynamic Parameter Variations}

The generalization capability evaluation first examined performance under modified environmental dynamics, including doubled peristaltic force amplitude and altered tissue material properties. As shown in Table~\ref{tab:sr_ae_metrics}, JEDP-RL maintained a 95\% success rate compared to PPO-RL's 89\% with 7.2\% higher TL cost (41.6 mm vs. 38.8 mm). The average positioning error decreased from 2.34 mm to 1.89 mm, demonstrating improved navigation precision under dynamic environmental conditions.

\setlength\tabcolsep{0.2em}
\begin{table}[H]
\caption{Comparative experiment result under material property variations } 
\centering
\begin{tabular}{@{}lccccc}
\toprule
\textbf{Policy}        & \textbf{YM} & \textbf{SR [\%] $\uparrow$} & \textbf{AE [mm] $\downarrow$} & \textbf{NS $\downarrow$} & \textbf{TL [mm] $\downarrow$} \\ \midrule
PPO-RL                     & 15000                   & 89              & 2.34    & 50.93  &      \textbf{38.8}         \\
JEDP-RL                    & 15000                    & \textbf{95}              & \textbf{1.89}   & \textbf{41.7}  &   41.6                    \\

PPO-RL                      & 10000                  & 72             & 2.45     & 72.0    &    52.3           \\
JEDP-RL                      & 10000                  & \textbf{98}       & \textbf{1.78}  & \textbf{37.2}   &    \textbf{34.0}             \\

PPO-RL                      & 25000                   & 87            & 2.36        & 50.0       &   \textbf{40.1}   \\ 
JEDP-RL              & 25000                   & \textbf{92}             & \textbf{2.15}      & \textbf{38.9}      &     44.12      \\
\bottomrule
\end{tabular}
\label{tab:sr_ae_metrics}
\end{table}

\begin{figure}[t!]
  \centering
  \includegraphics[width=0.90\linewidth]{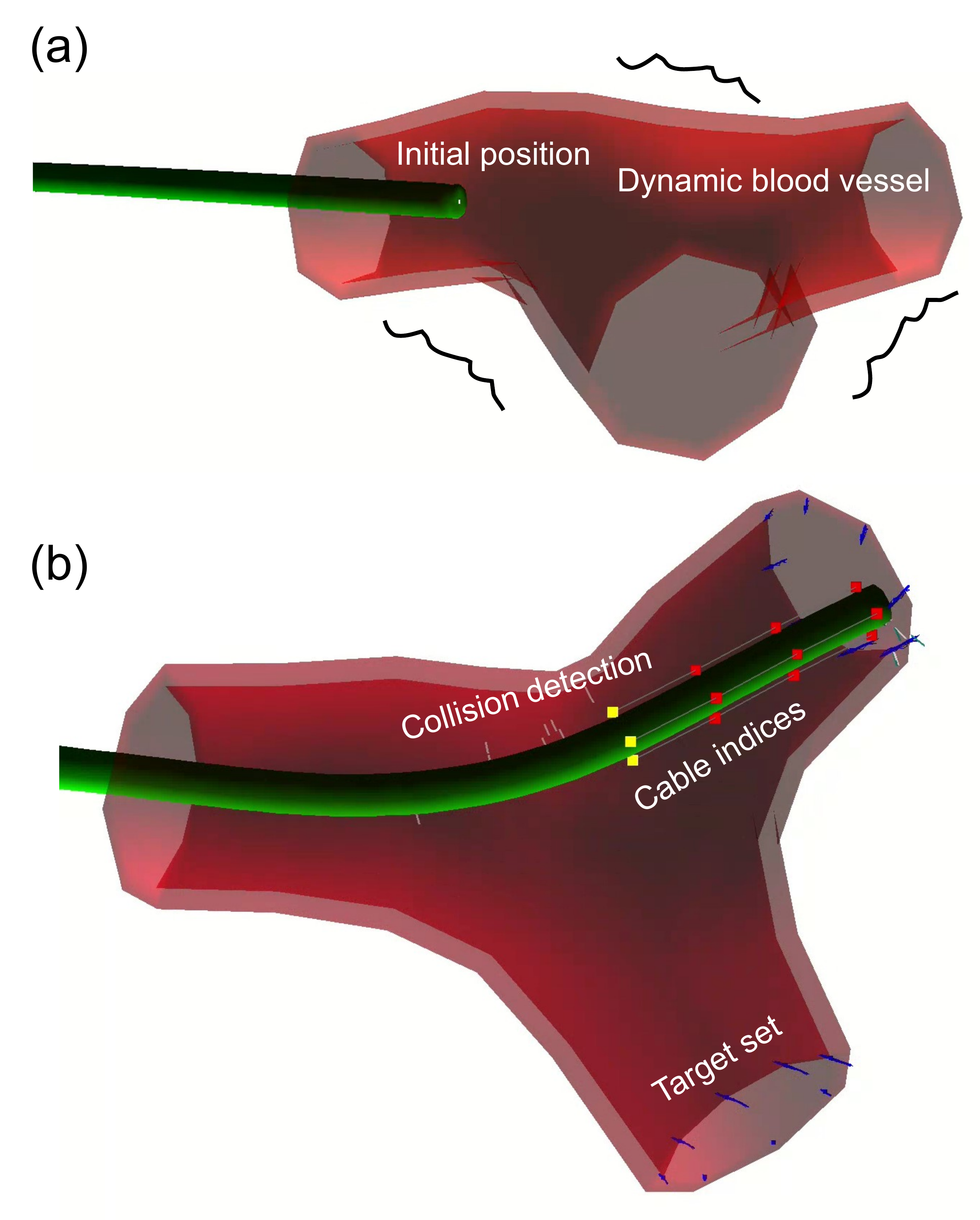}
  \caption{Blood vessel environment for evaluating pre-trained policies. We finetuned the policies in this environment. (a) The initial position of the robotic endoscope is shown. Periodic forces introduced dynamics. (b) Contact between the endoscope and blood vessel induces unmodeled deformation. The targets are randomly selected in the target set at the ends of two branches.}
  \label{Fig: evalenv}
\end{figure}

\subsubsection{Generalization to Robot Material Property Variations}
Experimental evaluation under varying material properties revealed JEDP-RL's superior adaptability to mechanical parameter deviations. When tested with substantially softer materials (YM=10,000), JEDP-RL demonstrated exceptional generalization capability, achieving a near-perfect 98\% success rate compared to PPO-RL's 72\%. This performance superiority extended across all metrics, showcasing 27.3\% lower average positioning error (1.78 mm vs. 2.45 mm), and 35\% shorter TL. This experimental validation aligns with theoretical expectations, where softer materials induce heightened structural unpredictability that critically undermines model-free approaches like PPO-RL lacking explicit state estimation mechanisms. By contrast, JEDP-RL's local exploration protocol effectively acquires Jacobian information, enabling precise deformation prediction and control compensation in compliant tissue environments.
\begin{figure*}[t!]
  \centering
  \includegraphics[width=0.95\linewidth]{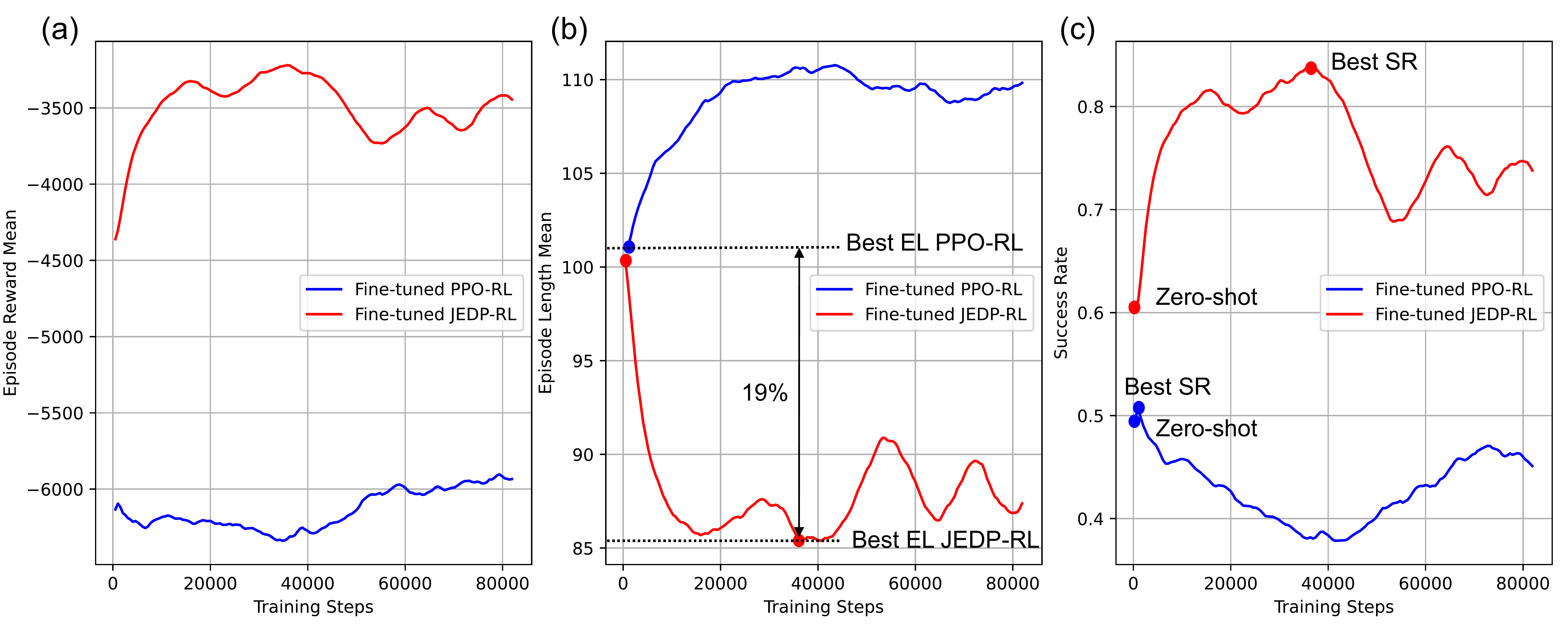}
  \caption{The policies of fine-tuned PPO-RL (blue) and the fine-tuned JEDP-RL (red). (a) Finetuning curves of episode reward means against training steps. (b) The best average number of episode lengths required for PPO-RL to reach the target is 19$\%$ higher compared to JEDP-RL. (c) Zero-shot performance of the policy trained by the JEDP-RL (red) is 11$\%$ higher than the PPO-RL (blue). The best SR obtained by the finetuned JEDP-RL is 83$\%$, which is 33$\%$ higher compared to finetuned PPO-RL.}
  \label{Fig: evaltensorboard}
\end{figure*}

Under stiff material conditions (YM=25,000), JEDP-RL maintained its performance advantage with a 92\% success rate versus 87\% for PPO-RL, coupled with an 8.9\% improvement in positioning accuracy (2.15 mm vs. 2.36 mm). While exhibiting a marginal 10\% trajectory cost increase in this regime, the associated 5\% success rate enhancement and 8.9\% error reduction justify this controlled efficiency trade-off, particularly considering clinical priorities favoring navigation reliability over minimal path optimization.

\subsubsection{Generalization to Unseen Environment}

The final generalization test evaluated policy performance in a fundamentally different anatomical environment: a simulated blood vessel with distinct biomechanical characteristics. As shown in Fig.~\ref{Fig: evalenv}, this environment introduced pulsatile flow forces and a challenging bifurcation geometry. 

Fig. \ref{Fig: evaltensorboard} presents the fine-tuning and evaluation results of both policies. As shown in Fig. \ref{Fig: evaltensorboard}(a), the episode reward mean during fine-tuning indicates a steady improvement for both policies, but JEDP-RL achieves a significantly higher reward compared to PPO-RL. This suggests that JEDP-RL adapts more efficiently to the new environment and learns effective navigation strategies in the presence of dynamic deformations. Fig. \ref{Fig: evaltensorboard}(b) shows that the best average episode length required for PPO-RL to reach the target is $19\%$ higher than that of JEDP-RL. This indicates that JEDP-RL achieves the target locations with more efficient trajectories, highlighting its advantage in adapting to environmental changes. Additionally, Fig. \ref{Fig: evaltensorboard}(c) compares the zero-shot performance of both policies before fine-tuning, showing that JEDP-RL (red) achieves $11\%$ higher SR compared to PPO-RL (blue). After fine-tuning, JEDP-RL further outperforms PPO-RL, achieving a maximum SR of $83\%$, which is $33\%$ higher than that of the fine-tuned PPO-RL.

The bifurcation navigation analysis reveals fundamental differences in policy adaptation strategies. While PPO-RL demonstrates branch-specific competence (Upper Branch: 82\% SR, Lower Branch: 7\% SR), JEDP-RL maintains balanced performance across both anatomical structures (UB: 82\% SR, LB: 69\% SR). This 9.8:1 success rate ratio disparity between branches for PPO-RL versus JEDP-RL's 1.2:1 ratio underscores the proposed method's capacity to develop generalized navigation strategies rather than environment-specific solutions.

\section{Conclusion}
\label{sec:conclusion}

This study establishes the efficacy of Jacobian exploratory dual-phase RL for deformable continuum robot (DCR) navigation in dynamic anatomical environments. The experimental results across stomach and vascular simulations demonstrate three fundamental advancements:

1). \textbf{Accelerated Policy Convergence}: The JEDP-RL framework achieved 3.2× faster convergence than PPO-RL (Fig. \ref{Fig: traintensorboard}(a)), validating the effectiveness of local Jacobian exploration. This acceleration stems from the dual-phase architecture that decouples mechanical parameter identification from policy optimization, effectively addressing the time-varying kinematics problem in DCRs.

2). \textbf{Enhanced Navigation Efficiency}: Post-convergence analysis revealed JEDP-RL requires 25\% fewer steps than PPO-RL to reach targets (Fig. \ref{Fig: traintensorboard}(b)).

3). \textbf{Superior Generalization Capability}: Under material property variations, JEDP-RL achieved over 92\% SR. In unseen vascular environments, JEDP-RL achieved an 83\% SR after finetuning versus PPO's 50\% (Fig. \ref{Fig: evaltensorboard}(c)).

While the suggested method yields significance, the increase in trajectory length caused by exploration remains a challenge that requires further attention.
For future research, we will further probe into learnable exploration action generation and relevant studies on sim2real algorithms.

\clearpage

\bibliography{IEEEabrv,ref}

\end{document}